# The Complexity of Reasoning with Cardinality Restrictions and Nominals in Expressive Description Logics

**Stephan Tobies**  TOBIES@INFORMATIK.RWTH-AACHEN.DE
*LuFG Theoretical Computer Science, RWTH Aachen*
*Ahornstr. 55, 52074 Aachen, Germany*

## Abstract

We study the complexity of the combination of the Description Logics $\mathcal{ALCQ}$ and $\mathcal{ALCQI}$ with a terminological formalism based on cardinality restrictions on concepts. These combinations can naturally be embedded into $C^2$, the two variable fragment of predicate logic with counting quantifiers, which yields decidability in NExpTime. We show that this approach leads to an optimal solution for $\mathcal{ALCQI}$, as $\mathcal{ALCQI}$ with cardinality restrictions has the same complexity as $C^2$ (NExpTime-complete). In contrast, we show that for $\mathcal{ALCQ}$, the problem can be solved in ExpTime. This result is obtained by a reduction of reasoning with cardinality restrictions to reasoning with the (in general weaker) terminological formalism of general axioms for $\mathcal{ALCQ}$ extended with nominals. Using the same reduction, we show that, for the extension of $\mathcal{ALCQI}$ with nominals, reasoning with general axioms is a NExpTime-complete problem. Finally, we sharpen this result and show that pure concept satisfiability for $\mathcal{ALCQI}$ with nominals is NExpTime-complete. Without nominals, this problem is known to be PSpace-complete.

## 1. Introduction

Description Logics (DLs) can be used in knowledge based systems to represent and reason about taxonomical knowledge of the application domain in a semantically well-defined manner (Woods & Schmolze, 1992). They allow the definition of complex concepts (i.e., classes, unary predicates) and roles (binary predicates) to be built from atomic ones by the application of a given set of constructors. For example, the following concept describes those parents having at least two daughters:

$$\texttt{Human} \sqcap (\texttt{Male} \sqcup \texttt{Female}) \sqcap (\geqslant 2\ \texttt{hasChild Female}) \sqcap \forall \texttt{hasChild.Human}$$

This concept is an example for the DL $\mathcal{ALCQ}$. $\mathcal{ALCQ}$ extends the "standard" DL $\mathcal{ALC}$ (Schmidt-Schauß & Smolka, 1991) by qualifying number restrictions, i.e., concepts restricting the number of individuals that are related via a given role (here hasChild), instead of allowing only for existential or universal restrictions like $\mathcal{ALC}$. $\mathcal{ALCQ}$ is a syntactic variant of the (multi-)modal logic K with *graded modalities* (Fine, 1972). In this paper we will study problems for the DLs $\mathcal{ALCQ}$ and $\mathcal{ALCQI}$. The latter extends $\mathcal{ALCQ}$ with the possibility to refer to the inverse of role relations. Additionally, in this paper we will encounter *nominals*, i.e., concepts referring to single elements of the domain. The extensions of $\mathcal{ALCQ}$ and $\mathcal{ALCQI}$ with nominals are denoted by $\mathcal{ALCQO}$ and $\mathcal{ALCQIO}$. An example concept of $\mathcal{ALCQIO}$ that describes the common children of the individuals ALICE and BOB living with ALICE or BOB is

$$\exists \texttt{hasChild}^{-1}.\texttt{ALICE} \sqcap \exists \texttt{hasChild}^{-1}.\texttt{BOB} \sqcap \exists \texttt{livesWith}.(\texttt{ALICE} \sqcup \texttt{BOB}).$$

TobiesTobies

|  | $\mathcal{ALCQ}$ | $\mathcal{ALCQO}$ | $\mathcal{ALCQI}$ | $\mathcal{ALCQIO}$ |
|---|---|---|---|---|
| Concept Satisfiability | PSpace-c. | open | PSpace-c. | **NExpTime**-c. |
| GCIs | ExpTime-c. | ExpTime-c. | ExpTime-c. | **NExpTime**-c. |
| Cardinality Restr. | **ExpTime**-c. | **ExpTime**-c. | **NExpTime**-c. | **NExpTime**-c. |

Figure 1: Complexity results established in this paper (shown in bold face)

Here, the parent relationship is expressed as the inverse of the `hasChild` relationship.

A terminological component (TBox) allows for the organisation of defined concepts and roles and forms the knowledge base of a DL system. TBoxes studied in DLs range from weak ones allowing only for the acyclic introduction of abbreviations for complex concepts, over TBoxes capable of expressing various forms of general axioms, to cardinality restrictions that can express restrictions on the number of elements the extension of a concept may have. In the following, we give examples of these three types of assertions.

The following TBox introduces parent as an abbreviation for a human having at least one child and whose children are all human, toddler for very young human, and busy parent for a parent having at least two children that are toddlers.

$$\texttt{Parent} = \texttt{Human} \sqcap (\geqslant 1 \texttt{ hasChild}) \sqcap \forall \texttt{hasChild.Human}$$
$$\texttt{Toddler} = \texttt{Human} \sqcap \texttt{VeryYoung}$$
$$\texttt{BusyParent} = \texttt{Parent} \sqcap (\geqslant 2 \texttt{ hasChild Toddler})$$

The next expressions are *general axioms* stating that males and females are disjoint ($\bot$ denotes the empty concept) and that males or females coincide with those humans having exactly two human parents.

$$\texttt{Female} \sqcap \texttt{Male} = \bot$$
$$\texttt{Female} \sqcup \texttt{Male} = \texttt{Human} \sqcap (= 2 \texttt{ hasChild}^{-1} \texttt{ Human})$$

Finally, the following expression is a *cardinality restriction* expressing that there are at most two earliest ancestors:

$$(\leq 2 \ (\texttt{Human} \sqcap (\leqslant 0 \texttt{ hasChild}^{-1} \texttt{ Human})))$$

Cardinality restriction were first introduced by Baader et al. (1996) as a terminological formalism for the DL $\mathcal{ALCQ}$; as we will see, they can express general axioms and hence are the most expressive of the terminological formalisms considered in this paper.

A key component of a DL system is a reasoning component that provides services like subsumption or consistency tests for the knowledge stored in the TBox. A subsumption test, for example, could infer from the previous definitions that both `Male` and `Female` are subsumed by `Human` and that `BusyParent` is subsumed by `Parent` as each busy parent must have at least one child. There exist sound and complete algorithms for reasoning in a large number of DLs and different TBox formalisms that are optimal with respect to the known worst-case complexity of these problems (see Donini et al., 1996, for an overview).

In this paper we establish a number of new complexity results for DLs with cardinality restrictions or nominals. Figure 1 summarises the new complexity bounds established in this paper. All problems are complete for their respective complexity class. This paper is organised as follows.

After giving some basic definitions in Section 2, we show that consistency of TBoxes with cardinality restrictions for $\mathcal{ALCQI}$ is a NExpTime-complete problem (Section 3). Membership in NExpTime is shown by a translation to the satisfiability problem of $C^2$ (Pacholski et al., 1997)[1], the two variable fragment of first order predicate logic augmented with counting quantifiers. The matching lower bound is established by a reduction from a NExpTime-complete bounded domino problem.

In Section 4, we show that reasoning with cardinality restrictions can be reduced to reasoning with the (weaker) formalism of general axioms in the presence of nominals. This yields interesting complexity results both for reasoning with cardinality restrictions and with nominals. Using a result from (De Giacomo, 1995), the reduction shows that consistency of TBoxes with cardinality restrictions for $\mathcal{ALCQ}$ is in ExpTime. This improves the result from (Baader et al., 1996), where it was shown that the problem can be solved in NExpTime. Moreover, we show that for a DL with number restrictions, inverse roles, and nominals reasoning problems become NExpTime-hard, which solves an open problem from (De Giacomo, 1995). This combination is of particular interest for the application of DLs in the area of reasoning with database schemata (Calvanese et al., 1998a, 1998b).

## 2. The Logic $\mathcal{ALCQI}$

**Definition 2.1** *Let $N_C$ be a set of atomic concept names and $N_R$ be a set of atomic role names. Concepts in $\mathcal{ALCQI}$ are built inductively from these using the following rules: all $A \in N_C$ are concepts, and, if $C$, $C_1$, and $C_2$ are concepts, then also*

$$\neg C, \; C_1 \sqcap C_2, \; \text{and} \; (\geqslant n \; S \; C),$$

*are concepts, where $n \in \mathbb{N}$ and $S = R$ or $S = R^{-1}$ for some $R \in N_R$.*

*A* cardinality restriction *of $\mathcal{ALCQI}$ is an expression of the form $(\geqslant n \; C)$ or $(\leqslant n \; C)$ where $C$ is a concept and $n \in \mathbb{N}$; an $\mathcal{ALCQI}$-$T_C$Box [2] is a finite set of cardinality restrictions.*

*The semantics of concepts is defined relative to an* interpretation $\mathcal{I} = (\Delta^{\mathcal{I}}, \cdot^{\mathcal{I}})$, *which consists of a domain $\Delta^{\mathcal{I}}$ and a valuation $(\cdot^{\mathcal{I}})$ that maps each concept name $A$ to a subset $A^{\mathcal{I}}$ of $\Delta^{\mathcal{I}}$ and each role name $R$ to a subset $R^{\mathcal{I}}$ of $\Delta^{\mathcal{I}} \times \Delta^{\mathcal{I}}$. This valuation is inductively extended to arbitrary concepts using the following rules, where $\sharp M$ denotes the cardinality of a set $M$:*

$$(\neg C)^{\mathcal{I}} := \Delta^{\mathcal{I}} \setminus C^{\mathcal{I}},$$
$$(C_1 \sqcap C_2)^{\mathcal{I}} := C_1^{\mathcal{I}} \cap C_2^{\mathcal{I}},$$
$$(\geqslant n \; R \; C)^{\mathcal{I}} := \{a \in \Delta^{\mathcal{I}} \mid \sharp\{b \in \Delta^{\mathcal{I}} \mid (a,b) \in R^{\mathcal{I}} \wedge b \in C^{\mathcal{I}}\} \geq n\},$$
$$(\geqslant n \; R^{-1} \; C)^{\mathcal{I}} := \{a \in \Delta^{\mathcal{I}} \mid \sharp\{b \in Δ^{\mathcal{I}} \mid (b,a) \in R^{\mathcal{I}} \wedge b \in C^{\mathcal{I}}\} \geq n\}.$$

---

1. The NExpTime-result is valid only if we assume unary coding of numbers in the counting quantifiers. This is the standard assumption made by most results concerning the complexity of DLs.
2. The subscripted "C" indicates that the TBox consists of cardinality restrictions

$$\begin{aligned}
\Psi_x(A) &:= Ax \quad &\text{for } A \in N_C \\
\Psi_x(\neg C) &:= \neg \Psi_x(C) \\
\Psi_x(C_1 \sqcap C_2) &:= \Psi_x(C_1) \wedge \Psi_x(C_2) \\
\Psi_x(\geqslant n\ R\ C) &:= \exists^{\geq n} y.(Rxy \wedge \Psi_y(C)) \\
\Psi_x(\geqslant n\ R^{-1}\ C) &:= \exists^{\geq n} y.(Ryx \wedge \Psi_y(C)) \\
\Psi_y(C) &:= \Psi_x(C)[x\backslash y, y\backslash x] \\
\Psi(\bowtie n\ C) &:= \exists^{\bowtie n} x.\Psi_x(C) \quad \text{for } \bowtie \in \{\geqslant, \leqslant\} \\
\Psi(T) &:= \bigwedge\{\Psi(\bowtie\ n\ C) \mid (\bowtie\ n\ C) \in T\}
\end{aligned}$$

Figure 2: The translation from $\mathcal{ALCQI}$ into $C^2$

An interpretation $\mathcal{I}$ satisfies *a cardinality restriction* $(\geqslant n\ C)$ iff $\sharp(C^{\mathcal{I}}) \geq n$, and it satisfies $(\leqslant n\ C)$ iff $\sharp(C^{\mathcal{I}}) \leq n$. It satisfies *a $T_C$Box $T$* iff it satisfies all cardinality restrictions in $T$; in this case, $\mathcal{I}$ is called a model *of $T$ and we will denote this fact by* $\mathcal{I} \models T$. A $T_C$Box that has a model is called consistent.

With $\mathcal{ALCQ}$ we denote the fragment of $\mathcal{ALCQI}$ that does not contain any inverse roles $R^{-1}$.

Using the constructors from Definition 2.1, we use $(\forall\ C)$ as an abbreviation for the cardinality restriction $(\leqslant 0\ \neg C)$ and introduce the following abbreviations for concepts:

$$\begin{aligned}
C_1 \sqcup C_2 &= \neg(\neg C_1 \sqcap \neg C_2) & (\leqslant n\ S\ C) &= \neg(\geqslant (n+1)\ S\ C) \\
C_1 \to C_2 &= \neg C_1 \sqcup C_2 & (= n\ S\ C) &= (\leqslant n\ S\ C) \sqcap (\geqslant n\ S\ C) \\
\exists S.C &= (\geqslant 1\ S\ C) & \top &= A \sqcup \neg A \quad \text{for some } A \in N_C \\
\forall S.C &= (\leqslant 0\ S\ \neg C)
\end{aligned}$$

TBoxes consisting of cardinality restrictions have first been studied in (Baader et al., 1996) for the DL $\mathcal{ALCQ}$. Obviously, two concepts $C, D$ have the same extension in an interpretation $\mathcal{I}$ iff $\mathcal{I}$ satisfies the cardinality restriction $(\leqslant 0\ (C \sqcap \neg D) \sqcup (\neg C \sqcap D))$. Hence, cardinality restrictions can express terminological axioms of the form $C = D$, which are satisfied by an interpretation $\mathcal{I}$ iff $C^{\mathcal{I}} = D^{\mathcal{I}}$. General axioms are the most expressive TBox formalisms usually studied in the DL context (De Giacomo & Lenzerini, 1996). One standard inference service for DL systems is *satisfiability* of a concept $C$ with respect to a $T_C$Box $T$, i.e., is there an interpretation $\mathcal{I}$ such that $\mathcal{I} \models T$ and $C^{\mathcal{I}} \neq \emptyset$. For a TBox formalism based on cardinality restrictions this is easily reduced to TBox consistency, because obviously $C$ is satisfiable with respect to $T$ iff $T \cup \{(\geqslant 1\ C)\}$ is a consistent $T_C$Box. For this the reason, we will restrict our attention to $T_C$Box consistency; other standard inferences such as concept subsumption can be reduced to consistency as well.

Until now there does not exist a direct decision procedure for $\mathcal{ALCQI}$ $T_C$Box consistency. Nevertheless this problem can be decided with the help of a well-known translation of $\mathcal{ALCQI}$-$T_C$Boxes to $C^2$ (Borgida, 1996), given in Figure 2. The logic $C^2$ is the fragment of predicate logic in which formulae may contain at most two variables, but which is enriched with counting quantifiers of the form $\exists^{\geq \ell}$. The translation $\Psi$ yields a satisfiable sentence of $C^2$ if and only if the translated $T_C$Box is consistent. Since the translation from $\mathcal{ALCQI}$ to $C^2$ can be performed in linear time, the NExpTime upper bound (Grädel et al., 1997; Pacholski et al., 1997) for satisfiability of $C^2$ directly carries over to $\mathcal{ALCQI}$-T$_C$Box consistency:

**Lemma 2.2** *Consistency of an $\mathcal{ALCQI}$-T$_C$Box $T$ can be decided in* NExpTime.

Please note that the NExpTime-completeness result from (Pacholski et al., 1997) is only valid if we assume unary coding of numbers in the input; this implies that a number $n$ may not be stored in logarithmic space in some $k$-ary representation but consumes $n$ units of storage. This is the standard assumption made by most results concerning the complexity of DLs. We will come back to this issue in Section 3.3.

## 3. Consistency of $\mathcal{ALCQI}$-T$_C$Boxes is NExpTime-complete

To show that NExpTime is also the lower bound for the complexity of T$_C$Box consistency, we use a bounded version of the domino problem. Domino problems (Wang, 1963; Berger, 1966) have successfully been employed to establish undecidability and complexity results for various description and modal logics (Spaan, 1993; Baader & Sattler, 1999).

### 3.1 Domino Systems

**Definition 3.1** *For $n \in \mathbb{N}$, let $\mathbb{Z}_n$ denote the set $\{0, \ldots, n-1\}$ and $\oplus_n$ denote the addition modulo $n$. A domino system is a triple $\mathcal{D} = (D, H, V)$, where $D$ is a finite set (of tiles) and $H, V \subseteq D \times D$ are relations expressing horizontal and vertical compatibility constraints between the tiles. For $s, t \in \mathbb{N}$, let $U(s,t)$ be the torus $\mathbb{Z}_s \times \mathbb{Z}_t$, and let $w = w_0 \ldots w_{n-1}$ be a word over $D$ of length $n$ (with $n \leq s$). We say that $\mathcal{D}$ tiles $U(s,t)$ with initial condition $w$ iff there exists a mapping $\tau : U(s,t) \to D$ such that, for all $(x,y) \in U(s,t)$,*

- *if $\tau(x,y) = d$ and $\tau(x \oplus_s 1, y) = d'$, then $(d,d') \in H$ (horizontal constraint);*
- *if $\tau(x,y) = d$ and $\tau(x, y \oplus_t 1) = d'$, then $(d,d') \in V$ (vertical constraint);*
- *$\tau(i,0) = w_i$ for $0 \leq i < n$ (initial condition).*

Bounded domino systems are capable of expressing the computational behaviour of restricted, so-called *simple*, Turing Machines (TM). This restriction is non-essential in the following sense: Every language accepted in time $T(n)$ and space $S(n)$ by some one-tape TM is accepted within the same time and space bounds by a simple TM, as long as $S(n), T(n) \geq 2n$ (Börger et al., 1997).

**Theorem 3.2 ((Börger et al., 1997), Theorem 6.1.2)**
Let $M$ be a simple TM with input alphabet $\Sigma$. Then there exists a domino system $\mathcal{D} = (D, H, V)$ and a linear time reduction which takes any input $x \in \Sigma^*$ to a word $w \in D^*$ with $|x| = |w|$ such that

- If $M$ accepts $x$ in time $t_0$ with space $s_0$, then $\mathcal{D}$ tiles $U(s,t)$ with initial condition $w$ for all $s \geq s_0 + 2, t \geq t_0 + 2$;
- if $M$ does not accept $x$, then $\mathcal{D}$ does not tile $U(s,t)$ with initial condition $w$ for any $s, t \geq 2$.

**Corollary 3.3**
There is a domino system $\mathcal{D}$ such that the following is a NExpTime-hard problem:

> Given an initial condition $w = w_0 \ldots w_{n-1}$ of length $n$. Does $\mathcal{D}$ tile the torus $U(2^{n+1}, 2^{n+1})$ with initial condition $w$?

**Proof.** Let $M$ be a (w.l.o.g. simple) non-deterministic TM with time- (and hence space-) bound $2^n$ deciding an arbitrary NExpTime-complete language $\mathcal{L}(M)$ over the alphabet $\Sigma$. Let $\mathcal{D}$ be the according domino system and *trans* the reduction from Theorem 3.2.

The function *trans* is a linear reduction from $\mathcal{L}(M)$ to the problem above: For $v \in \Sigma^*$ with $|v| = n$, it holds that $v \in \mathcal{L}(M)$ iff $M$ accepts $v$ in time and space $2^{|v|}$ iff $\mathcal{D}$ tiles $U(2^{n+1}, 2^{n+1})$ with initial condition $trans(v)$. ∎

### 3.2 Defining a Torus of Exponential Size

Similar to proving undecidability by reduction of unbounded domino problems, where defining infinite grids is the key problem, defining a torus of exponential size is the key to obtaining a NExpTime-completeness proof by reduction of bounded domino problems.

To be able to apply Corollary 3.3 to $T_C$Box consistency for $\mathcal{ALCQI}$, we must characterise the torus $\mathbb{Z}_{2^n} \times \mathbb{Z}_{2^n}$ with a $T_C$Box of polynomial size. To characterise this torus, we use $2n$ concepts $X_0, \ldots, X_{n-1}$ and $Y_0, \ldots, Y_{n-1}$, where $X_i$ (resp., $Y_i$) codes the $i$th bit of the binary representation of the X-coordinate (resp., Y-coordinate) of an element $a$.

For an interpretation $\mathcal{I}$ and an element $a \in \Delta^{\mathcal{I}}$, we define $pos(a)$ by

$$pos(a) := (xpos(a), ypos(a)) := \left( \sum_{i=0}^{n-1} x_i \cdot 2^i, \sum_{i=0}^{n-1} y_i \cdot 2^i \right), \text{ where}$$

$$x_i = \begin{cases} 0, & \text{if } a \notin X_i^{\mathcal{I}} \\ 1, & \text{otherwise} \end{cases} \qquad y_i = \begin{cases} 0, & \text{if } a \notin Y_i^{\mathcal{I}} \\ 1, & \text{otherwise .} \end{cases}$$

We use a well-known characterisation of binary addition (e.g. (Börger et al., 1997)) to relate the positions of the elements in the torus:

**Lemma 3.4** Let $x, x'$ be natural numbers with binary representations

$$x = \sum_{i=0}^{n-1} x_i \cdot 2^i \quad \text{and} \quad x' = \sum_{i=0}^{n-1} x'_i \cdot 2^i.$$

Then

$$x' \equiv x + 1 \pmod{2^n} \quad \text{iff} \quad \bigwedge_{k=0}^{n-1} (\bigwedge_{j=0}^{k-1} x_j = 1) \to (x_k = 1 \leftrightarrow x'_k = 0)$$

$$\wedge \bigwedge_{k=0}^{n-1} (\bigvee_{j=0}^{k-1} x_j = 0) \to (x_k = x'_k) \, ,$$

where the empty conjunction and disjunction are interpreted as true and false, respectively.

$$\begin{aligned}
T_n = \{ & (\forall\ \exists east.\top), & (\forall\ \exists north.\top), \\
& (\forall\ (=1\ east^{-1}\ \top)), & (\forall\ (=1\ north^{-1}\ \top)), \\
& (\geqslant 1\ C_{(0,0)}), & (\geqslant 1\ C_{(2^n-1,2^n-1)}), \\
& (\leqslant 1\ C_{(2^n-1,2^n-1)}), & (\forall\ D_{east} \sqcap D_{north})\ \} \\
C_{(0,0)} = & \prod_{k=0}^{n-1} \neg X_k \sqcap \prod_{k=0}^{n-1} \neg Y_k \\
C_{(2^n-1,2^n-1)} = & \prod_{k=0}^{n-1} X_k \sqcap \prod_{k=0}^{n-1} Y_k \\
D_{east} = & \prod_{k=0}^{n-1} (\prod_{j=0}^{k-1} X_j) \to ((X_k \to \forall east.\neg X_k) \sqcap (\neg X_k \to \forall east.X_k)) \sqcap \\
& \prod_{k=0}^{n-1} (\bigsqcup_{j=0}^{k-1} \neg X_j) \to ((X_k \to \forall east.X_k) \sqcap (\neg X_k \to \forall east.\neg X_k)) \sqcap \\
& \prod_{k=0}^{n-1} ((Y_k \to \forall east.Y_k) \sqcap (\neg Y_k \to \forall east.\neg Y_k)) \\
D_{north} = & \ldots
\end{aligned}$$

Figure 3: A T$_C$Box defining a torus of exponential size

The T$_C$Box $T_n$ is defined in Figure 3. The concept $C_{(0,0)}$ is satisfied by all elements $a$ of the domain for which $pos(a) = (0,0)$ holds. $C_{(2^n-1,2^n-1)}$ is a similar concept, whose instances $a$ satisfy $pos(a) = (2^n-1, 2^n-1)$.

The concept $D_{north}$ is similar to $D_{east}$ where the role *north* has been substituted for *east* and variables $X_i$ and $Y_i$ have been swapped. The concept $D_{east}$ (resp. $D_{north}$) enforces that, along the role *east* (resp. *north*), the value of *xpos* (resp. *ypos*) increases by one while the value of *ypos* (resp. *xpos*) is unchanged. They are analogous to the formula in Lemma 3.4.

The following lemma is a consequence of the definition of *pos* and Lemma 3.4.

**Lemma 3.5** Let $\mathcal{I} = (\Delta^\mathcal{I}, \cdot^\mathcal{I})$ be an interpretation, $D_{east}, D_{north}$ defined as in Figure 3, and $a, b \in \Delta^\mathcal{I}$.

$$\begin{aligned}
(a,b) \in east^\mathcal{I} \text{ and } a \in D_{east}^\mathcal{I} \text{ implies:} \quad & xpos(b) \equiv xpos(a) + 1 \pmod{2^n} \\
& ypos(b) = ypos(a) \\
(a,b) \in north^\mathcal{I} \text{ and } a \in D_{north}^\mathcal{I} \text{ implies:} \quad & xpos(b) = xpos(a) \\
& ypos(b) \equiv ypos(a) + 1 \pmod{2^n}
\end{aligned}$$

The T$_C$Box $T_n$ defines a torus of exponential size in the following sense:

**Lemma 3.6** Let $T_n$ be the T$_C$Box as defined in Figure 3. Let $\mathcal{I} = (\Delta^\mathcal{I}, \cdot^\mathcal{I})$ be a model of $T_n$. Then

$$(\Delta^\mathcal{I}, east^\mathcal{I}, north^\mathcal{I}) \cong (U(2^n, 2^n), S_1, S_2),$$

where $U(2^n, 2^n)$ is the torus $\mathbb{Z}_{2^n} \times \mathbb{Z}_{2^n}$ and $S_1, S_2$ are the horizontal and vertical successor relations on this torus.

**Proof.** We show that the function *pos* is an isomorphism from $\Delta^\mathcal{I}$ to $U(2^n, 2^n)$. Injectivity of *pos* is shown by induction on the "Manhattan distance" $d(a)$ of the *pos*-value of an element $a$ to the *pos*-value of the upper right corner.

For an element $a \in \Delta^\mathcal{I}$ we define $d(a)$ by

$$d(a) = (2^n - 1 - xpos(a)) + (2^n - 1 - ypos(a)).$$

Note that $pos(a) = pos(b)$ implies $d(a) = d(b)$. Since $\mathcal{I} \models (\leqslant 1\ C_{(2^n-1,2^n-1)})$, there is at most one element $a \in \Delta^\mathcal{I}$ such that $d(a) = 0$. Hence, there is exactly one element $a$ such that $pos(a) = (2^n - 1, 2^n - 1)$. Now assume there are elements $a, b \in \Delta^\mathcal{I}$ such that $pos(a) = pos(b)$ and $d(a) = d(b) > 0$. Then either $xpos(a) < 2^n - 1$ or $ypos(a) < 2^n - 1$. W.l.o.g., we assume $xpos(a) < 2^n - 1$. From $\mathcal{I} \models T_n$, it follows that $a, b \in (\exists east.\top)^\mathcal{I}$. Let $a_1, b_1$ be elements such that $(a, a_1) \in east^\mathcal{I}$ and $(b, b_1) \in east^\mathcal{I}$. Since $d(a_1) = d(b_1) < d(a)$ and $pos(a_1) = pos(b_1)$, the induction hypothesis yields $a_1 = b_1$. From Lemma 3.5 it follows that

$$xpos(a_1) \equiv xpos(b_1) \equiv xpos(a) + 1 \pmod{2^n}$$
$$ypos(a_1) = ypos(b_1) = ypos(a)$$

This also implies $a = b$ because $a_1 \in (= 1\ east^{-1}.\top)^\mathcal{I}$ and $\{(a, a_1), (b, a_1)\} \subseteq east^\mathcal{I}$. Hence *pos* is injective.

To prove that *pos* is also *surjective* we use a similar technique. This time, we use an induction on the distance from the lower left corner. For each element $(x, y) \in U(2^n, 2^n)$, we define:

$$d'(x, y) = x + y.$$

We show by induction that, for each $(x, y) \in U(2^n, 2^n)$, there is an element $a \in \Delta^\mathcal{I}$ such that $pos(a) = (x, y)$. If $d'(x, y) = 0$, then $x = y = 0$. Since $\mathcal{I} \models (\geqslant 1\ C_{(0,0)})$, there is an element $a \in \Delta^\mathcal{I}$ such that $pos(a) = (0, 0)$. Now consider $(x, y) \in U(2^n, 2^n)$ with $d'(x, y) > 0$. Without loss of generality we assume $x > 0$ (if $x = 0$ then $y > 0$ must hold). Hence $(x - 1, y) \in U(2^n, 2^n)$ and $d'(x - 1, y) < d'(x, y)$. From the induction hypothesis, it follows that there is an element $a \in \Delta^\mathcal{I}$ such that $pos(a) = (x - 1, y)$. Then there must be an element $a_1$ such that $(a, a_1) \in east^\mathcal{I}$ and Lemma 3.5 implies that $pos(a_1) = (x, y)$. Hence *pos* is also surjective.

Finally, *pos* is indeed a homomorphism as an immediate consequence of Lemma 3.5. ∎

It is interesting to note that we need inverse roles only to guarantee that *pos* is injective. The same can be achieved by adding the cardinality restriction $(\leqslant (2^n \cdot 2^n)\ \top)$ to $T_n$, from which the injectivity of *pos* follows from its surjectivity and simple cardinality considerations. Of course the size of this cardinality restriction would only be polynomial in $n$ if we assume binary coding of numbers. Also note that we have made explicit use of the special expressive power of cardinality restrictions by stating that, in any model of $T_n$, the extension of $C_{(2^n-1,2^n-1)}$ must have *at most* one element. This cannot be expressed with a $\mathcal{ALCQI}$-TBox consisting of terminological axioms.

### 3.3 Reducing Domino Problems to T$_C$Box Consistency

Once Lemma 3.6 has been proved, it is easy to reduce the bounded domino problem to T$_C$Box consistency. We use the standard reduction that has been applied in the DL context, e.g., in (Baader & Sattler, 1999).

**Lemma 3.7** Let $\mathcal{D} = (D, V, H)$ be a domino system. Let $w = w_0 \ldots w_{n-1} \in D^*$. There is a T$_C$Box $T(n, \mathcal{D}, w)$ such that:

- $T(n, \mathcal{D}, w)$ is consistent iff $\mathcal{D}$ tiles $U(2^n, 2^n)$ with initial condition $w$.
- $T(n, \mathcal{D}, w)$ can be computed in time polynomial in $n$.

**Proof.** We define $T(n, \mathcal{D}, w) := T_n \cup T_\mathcal{D} \cup T_w$, where $T_n$ is defined in Figure 3, $T_\mathcal{D}$ captures the vertical and horizontal compatibility constraints of the domino system $\mathcal{D}$, and $T_w$ enforces the initial condition. We use an atomic concept $C_d$ for each tile $d \in D$. $T_\mathcal{D}$ consists of the following cardinality restrictions:

$$(\forall \bigsqcup_{d \in D} C_d), \quad (\forall \bigsqcap_{d \in D} \bigsqcap_{d' \in D \setminus \{d\}} \neg (C_d \sqcap C_{d'})),$$

$$(\forall \bigsqcap_{d \in D} (C_d \to (\forall east. \bigsqcup_{(d,d') \in H} C_{d'}))), \quad (\forall \bigsqcap_{d \in D} (C_d \to (\forall north. \bigsqcup_{(d,d') \in V} C_{d'}))).$$

$\mathcal{T}_w$ consists of the cardinality restrictions

$$(\forall (C_{(0,0)} \to C_{w_0})), \ldots, (\forall (C_{(n-1,0)} \to C_{w_{n-1}})),$$

where, for each $x, y$, $C_{(x,y)}$ is a concept that is satisfied by an element $a$ iff $pos(a) = (x, y)$, defined similarly to $C_{(0,0)}$ and $C_{(2^n-1, 2^n-1)}$.

From the definition of $T(n, \mathcal{D}, w)$ and Lemma 3.6, it follows that each model of $T(n, \mathcal{D}, w)$ immediately induces a tiling of $U(2^n, 2^n)$ and vice versa. Also, for a fixed domino system $\mathcal{D}$, $T(n, \mathcal{D}, w)$ is obviously polynomially computable. ∎

The main result of this section is now an immediate consequence of Lemma 2.2, Lemma 3.7, and Corollary 3.3:

**Theorem 3.8**
Consistency of $\mathcal{ALCQI}$-T$_C$Boxes is NExpTime-complete, even if unary coding of numbers is used in the input.

Recalling the note below the proof of Lemma 3.6, we see that the same argument also applies to $\mathcal{ALCQ}$ if we allow binary coding of numbers.

**Corollary 3.9**
Consistency of $\mathcal{ALCQ}$-T$_C$Boxes is NExpTime-hard, if binary coding is used to represent numbers in cardinality restrictions.

It should be noted that it is open if the problem can be decided in NExpTime, if binary coding of numbers is used, since the reduction of $C^2$ only yields decidability in 2-NExpTime.

In the following section, we will see that, for unary coding of numbers, deciding consistency of $\mathcal{ALCQ}$-$T_C$Boxes can be done in ExpTime (Corollary 4.8). This shows that the coding of numbers indeed has an influence on the complexity of the reasoning problem. It is worth noting that the complexity of pure concept satisfiability for $\mathcal{ALCQ}$ does not depend on the coding; the problem is PSpace-complete both for binary and unary coding of numbers (Tobies, 2000).

For unary coding, we needed both inverse roles and cardinality restrictions for the reduction. This is consistent with the fact that satisfiability for $\mathcal{ALCQI}$ concepts with respect to TBoxes consisting of terminological axioms is still in ExpTime, which can be shown by a reduction to the ExpTime-complete logics $\mathcal{CIN}$ (De Giacomo, 1995) or CPDL (Pratt, 1979). This shows that cardinality restrictions on concepts are an additional source of complexity. One reason for this might be that $\mathcal{ALCQI}$ with cardinality restrictions no longer has the tree-model property.

## 4. Reasoning with Nominals

Nominals, i.e., atomic concepts referring to single individuals of the domain, are studied both in the area of DLs (Borgida & Patel-Schneider, 1994; Donini et al., 1996) and modal logics (Gargov & Goranko, 1993; Blackburn & Seligman, 1996; Areces et al., 1999). In this section we show how, in the presence of nominals, consistency for $T_C$Boxes can be polynomially reduced to consistency of TBoxes consisting of general inclusion axioms, which, in general, is an easier problem. This correspondence is used to obtain two novel results: (i) we show that, for unary coding, consistency of $\mathcal{ALCQ}$-TBoxes consisting of cardinality restrictions can be decided in ExpTime; (ii) we show that, in the presence of both inverse roles and number restrictions, reasoning with nominals is strictly harder than without nominals: the complexity of determining consistency of TBoxes with general axioms rises from ExpTime to NExpTime, and the complexity of concept satisfiability rises from PSpace to NExpTime.

**Definition 4.1** *Let $N_I$ be a set of* individual names *(also called* nominals*) disjoint from $N_C$ and $N_R$. Concepts in $\mathcal{ALCQIO}$ are defined as $\mathcal{ALCQI}$-concepts with the additional rule that, for every $o \in N_I$, $o$ is an $\mathcal{ALCQIO}$-concept.*

*A* general concept inclusion axiom *for $\mathcal{ALCQIO}$ is an expression of the from $C \sqsubseteq D$, where $C$ and $D$ are $\mathcal{ALCQIO}$-concepts. A $T_I$Box for $\mathcal{ALCQIO}$ is a finite set of general inclusion axioms for $\mathcal{ALCQIO}$, where the subscript "I" stands for "Inclusion".*

*The semantics of $\mathcal{ALCQIO}$ concepts is defined similar as for $\mathcal{ALCQI}$, with the additional requirement that every interpretation maps a nominal $o \in N_I$ to a singleton set $o^{\mathcal{I}} \subseteq \Delta^{\mathcal{I}}$; $o$ can be seen as a name for the element in $o^{\mathcal{I}}$. Please note that we do not have a* unique name assumption, *i.e., we do not assume that $o_1 \neq o_2$ implies $o_1^{\mathcal{I}} \neq o_2^{\mathcal{I}}$.*

*An interpretation $\mathcal{I}$ satisfies an axiom $C \sqsubseteq D$ iff $C^{\mathcal{I}} \subseteq D^{\mathcal{I}}$. It satisfies a $T_I$Box $T_{gci}$ iff it satisfies all axioms in $T_{gci}$; in this case $\mathcal{I}$ is called a* model *of $T_{gci}$, and we will denote this fact by $\mathcal{I} \models T_{gci}$. A $T_I$Box that has a model is called* consistent.

*Cardinality restrictions, $T_C$Boxes, and their interpretation for $\mathcal{ALCQIO}$ are defined analogously to $\mathcal{ALCQI}$.*

With $\mathcal{ALCQO}$ we denote the fragment of $\mathcal{ALCQIO}$ that does not contain any inverse roles $R^{-1}$.

**Lemma 4.2** Consistency of $T_C$Boxes or $T_I$Boxes both for $\mathcal{ALCQO}$ and $\mathcal{ALCQIO}$ is ExpTime-hard and can be decided in NExpTime, if unary coding of numbers is used.

**Proof.** Consistency of $T_I$Boxes (and hence of $T_C$Boxes) is ExpTime-hard already for (a syntactical variant of) $\mathcal{ALC}$ (Halpern & Moses, 1992). Assuming unary coding of numbers, we can reduce the problems to satisfiability of $C^2$, which yields the NExpTime upper bound. ∎

### 4.1 Expressing Cardinality Restrictions Using Nominals

In the following we show how, under the assumption of unary coding of numbers, consistency of $\mathcal{ALCQI}$-$T_C$Boxes can be polynomially reduced to consistency of $\mathcal{ALCQIO}$-$T_I$Boxes. It should be noted that, conversely, it is also possible to polynomially reduce consistency of $\mathcal{ALCQIO}$-$T_I$Boxes to consistency of $\mathcal{ALCQI}$-$T_C$Boxes: for an arbitrary concept $C$, the cardinality restrictions $\{(\leqslant 1\ C), (\geqslant 1\ C)\}$ force the interpretation of $C$ to be a singleton. Since we do not gain any further insight from this reduction, we do not formally prove this result.

**Definition 4.3** Let $T = \{(\bowtie_1\ n_1\ C_1), \ldots (\bowtie_k\ n_k\ C_k)\}$ be an $\mathcal{ALCQI}$-$T_C$Box. W.l.o.g., we assume that $T$ contains no cardinality restriction of the form $(\geqslant 0\ C)$ as these are trivially satisfied by any interpretation. The translation of $T$, denoted by $\Phi(T)$, is the $\mathcal{ALCQIO}$-$T_I$Box defined as follows:

$$\Phi(T) = \bigcup \{\Phi(\bowtie_i\ n_i\ C_i) \mid 1 \leq i \leq k\},$$

where $\Phi(\bowtie_i\ n_i\ C_i)$ is defined depending on whether $\bowtie_i = \leqslant$ or $\bowtie_i = \geqslant$.

$$\Phi(\bowtie_i\ n_i\ C_i) = \begin{cases} \{C_i \sqsubseteq o_i^1 \sqcup \cdots \sqcup o_i^{n_i}\} & \text{if } \bowtie_i = \leqslant \\ \{o_i^j \sqsubseteq C_i \mid 1 \leq j \leq n_i\} \cup \{o_i^j \sqsubseteq \neg o_i^\ell \mid 1 \leq j < \ell \leq n_i\} & \text{if } \bowtie_i = \geqslant \end{cases},$$

where $o_i^1, \ldots, o_i^{n_i}$ are fresh and distinct nominals and we use the convention that the empty disjunction is interpreted as $\neg \top$ to deal with the case $n_i = 0$.

Assuming unary coding of numbers, the translation of a $T_C$Box $T$ is obviously computable in polynomial time.

**Lemma 4.4** Let $T$ be an $\mathcal{ALCQI}$-$T_C$Box. $T$ is consistent iff $\Phi(T)$ is consistent.

**Proof.** Let $T = \{(\bowtie_1\ n_1\ C_1), \ldots (\bowtie_k\ n_k\ C_k)\}$ be a consistent $T_C$Box. Hence, there is a model $\mathcal{I}$ of $T$, and $\mathcal{I} \models (\bowtie_i\ n_i\ C_i)$ for each $1 \leq i \leq k$. We show how to construct a model $\mathcal{I}'$ of $\Phi(T)$ from $\mathcal{I}$. $\mathcal{I}'$ will be identical to $\mathcal{I}$ in every respect except for the interpretation of the nominals $o_i^j$ (which do not appear in $T$).

If $\bowtie_i = \leqslant$, then $\mathcal{I} \models T$ implies $\sharp C_i^\mathcal{I} \leq n_i$. If $n_i = 0$, then we have not introduced new nominals, and $\Phi(T)$ contains $C_i \sqsubseteq \neg \top$. Otherwise, we define $(o_i^j)^{\mathcal{I}'}$ such that $C_i^\mathcal{I} \subseteq$ $\{(o_i^j)^{\mathcal{I}'} \mid 1 \leq j \leq n_i\}$. This implies $C_i^{\mathcal{I}'} \subseteq (o_i^1)^{\mathcal{I}'} \cup \cdots \cup (o_i^{n_i})^{\mathcal{I}'}$ and hence, in either case, $\mathcal{I}' \models \Phi(\leqslant n_i\ C_i)$.

If $\bowtie_i = \geqslant$, then $n_i > 0$ must hold, and $\mathcal{I} \models T$ implies $\sharp C_i^{\mathcal{I}} \geq n_i$. Let $x_1, \ldots x_{n_i}$ be $n_i$ distinct elements from $\Delta^{\mathcal{I}}$ with $\{x_1, \ldots, x_{n_i}\} \subseteq C_i^{\mathcal{I}}$. We set $(o_i^j)^{\mathcal{I}'} = \{x_j\}$. Since we have chosen distinct individuals to interpret different nominals, we have $\mathcal{I}' \models o_i^j \sqsubseteq \neg o_i^\ell$ for every $1 \leq i < \ell \leq n_i$. Moreover, $x_j \in C_i^{\mathcal{I}}$ implies $\mathcal{I}' \models o_i^j \sqsubseteq C_i$ and hence $\mathcal{I}' \models \Phi(\geqslant n_i\ C_i)$.

We have chosen distinct nominals for every cardinality restrictions, hence the previous construction is well-defined and, since $\mathcal{I}'$ satisfies $\Phi(\bowtie_i n_i\ C_i)$ for every $i$, $\mathcal{I}' \models \Phi(T)$.

For the converse direction, let $\mathcal{I}$ be a models of $\Phi(T)$. The fact that $\mathcal{I} \models T$ (and hence the consistency of $T$) can be shown as follows: let $(\bowtie_i n_i\ C_i)$ be an arbitrary cardinality restriction in $T$. If $\bowtie_i = \leqslant$ and $n_i = 0$, then we have $\Phi(\leqslant 0\ C_i) = \{C_i \sqsubseteq \neg \top\}$ and, since $\mathcal{I} \models \Phi(T)$, we have $C_i^{\mathcal{I}} = \emptyset$ and hence $\mathcal{I} \models (\leqslant 0\ C_i)$. If $\bowtie_i = \leqslant$ and $n_i > 0$, we have $\{C_i \sqsubseteq o_i^1 \sqcup \cdots \sqcup o_i^{n_i}\} \subseteq \Phi(T)$. From $\mathcal{I} \models \Phi(T)$ follows $\sharp C_i^{\mathcal{I}} \leq \sharp(o_i^1 \sqcup \cdots \sqcup o_i^{n_i})^{\mathcal{I}} \leq n_i$. If $\bowtie_i = \geqslant$, then we have $\{o_i^j \sqsubseteq C_i \mid 1 \leq j \leq n_i\} \cup \{o_i^j \sqsubseteq \neg o_i^\ell \mid 1 \leq j < \ell \leq n_i\} \subseteq \Phi(T)$. From the first set of axioms we get $\{(o_i^j)^{\mathcal{I}} \mid 1 \leq j \leq n_i\} \subseteq C_i^{\mathcal{I}}$. From the second set of axioms we get that, for every $1 \leq j < \ell \leq n_i$, $(o_i^j)^{\mathcal{I}} \neq (o_i^\ell)^{\mathcal{I}}$. This implies that $n_i = \sharp \bigcup \{(o_i^j)^{\mathcal{I}} \mid 1 \leq j \leq n_i\} \leq \sharp C_i^{\mathcal{I}}$. ∎

**Theorem 4.5**
Assuming unary coding of numbers, consistency of $\mathcal{ALCQI}$-T$_C$Boxes can be polynomially reduced to consistency of $\mathcal{ALCQIO}$-T$_I$Boxes. Similarly, consistency of $\mathcal{ALCQ}$-T$_C$Boxes can be polynomially reduced to consistency of $\mathcal{ALCQO}$-T$_I$Boxes.

**Proof.** The first proposition follows from the fact that $\Phi(T)$ is polynomially computable from $T$ if we assume unary coding of numbers and from Lemma 4.4. The second proposition follows from the fact that the translation does not introduce additional inverse roles. If $T$ does not contain inverse roles, then neither does $\Phi(T)$, and hence the result of translating an $\mathcal{ALCQ}$-T$_C$Box is an $\mathcal{ALCQO}$-T$_I$Box. ∎

### 4.2 Complexity Results

We will now use Theorem 4.5 to obtain new complexity results both for DLs with cardinality restrictions and with nominals.

#### 4.2.1 $\mathcal{ALCQ}$ and $\mathcal{ALCQO}$

De Giacomo (1995) obtains complexity results for various DLs by sophisticated polynomial reduction to a propositional dynamic logic. The author establishes many complexity results, one of which is of special interest for our purposes.

**Theorem 4.6 ((De Giacomo, 1995), Section 7.3)**
Satisfiability and logical implication for $\mathcal{CNO}$ knowledge bases (TBox and ABox) are ExpTime-complete.

The DL $\mathcal{CNO}$ studied by the author is a strict extension of $\mathcal{ALCQO}$; TBoxes in his thesis correspond to what we call T$_I$Boxes in this paper. Unary coding of numbers is assumed throughout his thesis. Although a unique name assumption is made, it is not inherent to the utilised reduction since is explicitly enforced. It is thus possible to eliminate the propositions that require a unique interpretation of individuals from the reduction. Hence, together with Lemma 4.2, we get the following corollary.

**Corollary 4.7**
Consistency of $\mathcal{ALCQO}$-T$_I$Boxes is ExpTime-complete if unary coding of number is assumed.

Together with Theorem 4.5, this solves the open problem concerning the lower bound for the complexity of $\mathcal{ALCQ}$ with cardinality restrictions; moreover, it shows that the NExpTime-algorithm presented in (Baader et al., 1996) is not optimal with respect to worst case complexity.

**Corollary 4.8**
Consistency of $\mathcal{ALCQ}$-T$_C$Boxes is ExpTime-complete, if unary coding of numbers in cardinality and number restrictions is used.

### 4.2.2 $\mathcal{ALCQIO}$

Conversely, using Theorem 4.5 enables us to transfer the NExpTime-completeness result from Theorem 3.8 to $\mathcal{ALCQIO}$.

**Corollary 4.9**
Consistency of $\mathcal{ALCQIO}$-T$_I$Boxes or T$_C$Boxes is NExpTime-complete.

**Proof.** For T$_I$Boxes, this is an immediate corollary of Theorem 4.5 and Theorem 3.8. Reasoning with T$_C$Boxes is as hard as for T$_I$Boxes in the presences of nominals. ∎

This results explains a gap in (De Giacomo, 1995). There the author establishes the complexity of satisfiability of knowledge bases consisting of T$_I$Boxes and ABoxes both for $\mathcal{CNO}$, which allows for qualifying number restrictions, and for $\mathcal{CIO}$, which allows for inverse roles, by reduction to an ExpTime-complete PDL. No results are given for the combination $\mathcal{CINO}$, which is a strict extension of $\mathcal{ALCQIO}$. Corollary 4.8 shows that, assuming ExpTime $\neq$ NExpTime, there cannot be a polynomial reduction from the satisfiability problem of $\mathcal{CINO}$ knowledge bases to an ExpTime-complete PDL. Again, a possible explanation for this leap in complexity is the loss of the tree model property. While for $\mathcal{CIO}$ and $\mathcal{CNO}$, consistency is decided by searching for a tree-like pseudo-models even in the presence of nominals, this seems no longer to be possible in the case of knowledge bases for $\mathcal{CINO}$.

**Unique Name Assumption** It should be noted that our definition of nominals is non-standard for Description Logics in the sense that we do not impose the unique name assumption that is widely made, i.e., for any two individual names $o_1, o_2 \in N_I$, $o_1^\mathcal{I} \neq o_2^\mathcal{I}$ is required. Even without a unique name assumption, it is possible to enforce distinct interpretation of nominals by adding axioms of the form $o_1 \sqsubseteq \neg o_2$. Moreover, imposing a unique name assumption in the presence of inverse roles and number restriction leads to peculiar effects. Consider the following T$_I$Box:

$$T = \{o \sqsubseteq (\leqslant k\ R\ \top),\ \top \sqsubseteq \exists R^-.o\}$$

Under the unique name assumption, $T$ is consistent iff $N_I$ contains at most $k$ individual names, because each individual name must be interpreted by a unique element of the domain, every element of the domain must be reachable from $o^\mathcal{I}$ via the role $R$, and $o^\mathcal{I}$ may have at most $k$ $R$-successors. We believe that this dependency of the consistency of a $T_I$Box on constraints that are not explicit in the $T_I$Box is counter-intuitive and hence have not imposed the unique name assumption.

Nevertheless, it is possible to obtain a tight complexity bound for consistency of $\mathcal{ALCQIO}$-$T_I$Boxes with the unique name assumption without using Theorem 4.5, but by an immediate adaption of the proof of Theorem 3.8.

**Corollary 4.10**
Consistency of $\mathcal{ALCQIO}$-$T_I$Boxes *with the unique name assumption* is NExpTime-complete if unary coding of numbers assumed.

**Proof.** A simple inspection of the reduction used to prove Theorem 3.8, and especially of the proof of Lemma 3.6 shows that only a single nominal, which marks the upper right corner of the torus, is necessary to perform the reduction. If $o$ is an individual name and *create* is a role name, then the following $T_I$Box defines a torus of exponential size:

$$T_n = \{\ \top \sqsubseteq \exists east.\top, \quad \top \sqsubseteq \exists north.\top,$$
$$\top \sqsubseteq (= 1\ east^{-1}\ \top), \quad \top \sqsubseteq (= 1\ north^{-1}\ \top),$$
$$\top \sqsubseteq \exists create.C_{(0,0)}, \quad o \sqsubseteq C_{(2^n-1, 2^n-1)},$$
$$C_{(2^n-1, 2^n-1)} \sqsubseteq o, \quad \top \sqsubseteq D_{east} \sqcap D_{north}\ \}$$

Since this reduction uses only a single individual name, the unique name assumption is irrelevant. ∎

**Internalisation of Axioms** In the presence of inverse roles and nominals, it is possible to *internalise* general inclusion axioms into concepts using the *spy-point* technique used, e.g., in (Blackburn & Seligman, 1996; Areces et al., 1999). The main idea of this technique is to enforce that all elements in the model of a concept are reachable from a distinct point (the spy-point) marked by an individual name in a single step.

**Definition 4.11** *Let $T$ be an $\mathcal{ALCQIO}$-$T_I$Box. W.l.o.g., we assume that $T$ is of the form $\{\top \sqsubseteq C_1, \ldots, \top \sqsubseteq C_k\}$. Let spy denote a fresh role name and $i$ a fresh individual name. We define the function $\cdot^{spy}$ inductively on the structure of concepts by setting $A^{spy} = A$ for all $A \in N_C$, $o^{spy} = o$ for all $o \in N_I$, $(\neg C)^{spy} = \neg C^{spy}$, $(C_1 \sqcap C_2)^{spy} = C_1^{spy} \sqcap C_2^{spy}$, and $(\geqslant n\ R\ C)^{spy} = (\geqslant n\ R\ (\exists spy^-.i) \sqcap C^{spy})$.*

*The* internalisation *$C_T$ of $T$ is defined as follows:*

$$C_T = i \sqcap \bigsqcap_{\top \sqsubseteq C \in T} C^{spy} \sqcap \bigsqcap_{\top \sqsubseteq C \in T} \forall spy.C^{spy}$$

**Lemma 4.12** *Let $T$ be an $\mathcal{ALCQIO}$-$T_I$Box. $T$ is consistent iff $C_T$ is satisfiable.*

**Proof.** For the *if*-direction let $\mathcal{I}$ be a model of $C_T$ with $a \in (C_T)^\mathcal{I}$. This implies $i^\mathcal{I} = \{a\}$. Let $\mathcal{I}'$ be defined by

$$\Delta^{\mathcal{I}'} = \{a\} \cup \{x \in \Delta^\mathcal{I} \mid (a, x) \in spy^\mathcal{I}\}$$

and $\cdot^{\mathcal{I}'} = \cdot^{\mathcal{I}}|_{\Delta^{\mathcal{I}'}}$.

CLAIM 1: For every $x \in \Delta^{\mathcal{I}'}$ and every $\mathcal{ALCQIO}$-concept $C$, we have $x \in (C^{spy})^{\mathcal{I}}$ iff $x \in C^{\mathcal{I}'}$.

We proof this claim by induction on the structure of $C$. The only interesting case is $C = (\geqslant n\ R\ D)$. In this case $C^{spy} = (\geqslant n\ R\ (\exists spy^-.i) \sqcap D^{spy})$. We have

$$x \in (\geqslant n\ R\ (\exists spy^-.i) \sqcap D^{spy})^{\mathcal{I}}$$
$$\text{iff } \sharp\{y \in \Delta^{\mathcal{I}} \mid (x,y) \in R^{\mathcal{I}} \text{ and } y \in (\exists spy^-.i)^{\mathcal{I}} \cap (D^{spy})^{\mathcal{I}}\} \geqslant n$$
$$(*)\ \text{iff } \sharp\{y \in \Delta^{\mathcal{I}'} \mid (x,y) \in R^{\mathcal{I}'} \text{ and } y \in D^{\mathcal{I}'}\} \geqslant n$$
$$\text{iff } x \in (\geqslant n\ R\ D)^{\mathcal{I}'},$$

where the equivalence $(*)$ holds because of set equality and the definition of $\mathcal{I}'$.

By construction, for every $\top \sqsubseteq C \in T$ and every $x \in \Delta^{\mathcal{I}'}$, $x \in (C^{spy})^{\mathcal{I}}$. Due to Claim 1, this implies $x \in C^{\mathcal{I}'}$ and hence $\mathcal{I}' \models \top \sqsubseteq C$.

For the *only-if*-direction, let $\mathcal{I}$ be an interpretation with $\mathcal{I} \models T$. We pick an arbitrary element $a \in \Delta^{\mathcal{I}}$ and define an extension $\mathcal{I}'$ of $\mathcal{I}$ by setting $i^{\mathcal{I}'} = \{a\}$ and $spy^{\mathcal{I}'} = \{(a,x) \mid x \in \Delta^{\mathcal{I}}\}$. Since $i$ and $spy$ do not occur in $T$, we still have that $\mathcal{I}' \models T$.

CLAIM 2: For every $x \in \Delta^{\mathcal{I}'}$ and every $\mathcal{ALCQIO}$-concept $C$ that does not contain $i$ or $spy$, $x \in C^{\mathcal{I}'}$ iff $x \in (C^{spy})^{\mathcal{I}'}$.

Again, this claim is proved by induction on the structure of concepts and the only interesting case is $C = (\geqslant n\ R\ D)$.

$$x \in (\geqslant n\ R\ D)^{\mathcal{I}'}$$
$$\text{iff } \sharp\{y \in \Delta^{\mathcal{I}'} \mid (x,y) \in R^{\mathcal{I}'} \text{ and } y \in D^{\mathcal{I}'}\} \geqslant n$$
$$(*)\ \text{iff } \sharp\{y \in \Delta^{\mathcal{I}'} \mid (x,y) \in R^{\mathcal{I}'}, (a,y) \in spy^{\mathcal{I}'}, \text{ and } y \in (D^{spy})^{\mathcal{I}'}\} \geqslant n$$
$$\text{iff } x \in (\geqslant n\ R\ (\exists spy^-.i) \sqcap D^{spy})^{\mathcal{I}'}.$$

Again, the equivalence $(*)$ holds due to set equality and the definition of $\mathcal{I}'$.

Since, for every $\top \sqsubseteq C \in T$, we have $\mathcal{I}' \models \top \sqsubseteq C$, Claim 2 yields that $(\bigsqcap_{\top \sqsubseteq C \in T} C^{spy})^{\mathcal{I}'} = \Delta^{\mathcal{I}'}$ and hence $a \in (C_T)^{\mathcal{I}'}$ ∎

As a consequence, we obtain the sharper result that already pure concept satisfiability for $\mathcal{ALCQIO}$ is a NExpTime-complete problem.

**Corollary 4.13**
Concept satisfiability for $\mathcal{ALCQIO}$ is NExpTime-complete.

**Proof.** From Lemma 4.12, we get that the function mapping a $\mathcal{ALCQIO}$-T$_I$Box $T$ to $C_T$ is a reduction from consistency of $\mathcal{ALCQIO}$-T$_I$Boxes to $\mathcal{ALCQIO}$-concept satisfiability. From Corollary 4.9 we know that the former problem is NExpTime-complete. Obviously, $C_T$ can be computed from $T$ in polynomial time. Hence, the lower complexity bound transfers. ∎

|  | $\mathcal{ALCQ}$ | $\mathcal{ALCQO}$ | $\mathcal{ALCQI}$ | $\mathcal{ALCQIO}$ |
|---|---|---|---|---|
| Concept Satisfiability | PSpace-c. | open | PSpace-c. | NExpTime-c. |
| GCIs | ExpTime-c. | ExpTime-c. | ExpTime-c. | NExpTime-c. |
| Cardinality Restr. | ExpTime-c. | ExpTime-c. | NExpTime-c. | NExpTime-c. |

Figure 4: Complexity of the reasoning problems

## 5. Conclusion

Combining the results from (De Giacomo, 1995) and (Tobies, 2000) with the results from this paper, we obtain the classification of the complexity of concept satisfiability and TBox-consistency for various DLs and for TBoxes consisting either of cardinality restrictions or of general concept inclusion axioms shown in Figure 4, where we assume unary coding of numbers.

The result for $\mathcal{ALCQIO}$ shows that the current efforts of extending very expressive DLs as the logic $\mathcal{SHIQ}$ (Horrocks et al., 1999) and $\mathcal{DLR}$ (Calvanese et al., 1998c) or propositional dynamic logics as $\mathsf{CPDL}_g$ (De Giacomo & Lenzerini, 1996) with nominals are difficult tasks, if one wants to obtain a practical decision procedure, since already concept satisfiability for these logics is a NExpTime-hard problem.

We have shown that, while having the same complexity as $C^2$, $\mathcal{ALCQI}$ does not reach its expressive power (Tobies, 1999). Cardinality restrictions, although interesting for knowledge representation, are inherently hard to handle algorithmically. The same applies to nominals in the presence of inverse roles and number restrictions. As an explanation for this we offer the lack of a tree model property, which was identified by Vardi (1997) as an explanation for good algorithmic behaviour of many modal logics.

At a first glance, our results make $\mathcal{ALCQI}$ with cardinality restrictions on concepts or $\mathcal{ALCQIO}$ with general axioms obsolete for knowledge representation because $C^2$ delivers more expressive power at the same computational price. Yet, is is likely that a dedicated algorithm for $\mathcal{ALCQI}$ may have better average complexity than the $C^2$ algorithm; such an algorithm has yet to be developed. This is highly desirable as it would also be applicable to reasoning problems for expressive DLs with nominals, which have applications in the area of reasoning with database schemata (Calvanese et al., 1998a, 1998b).

An interesting question lies in the coding of numbers: If we allow binary coding of numbers, the translation approach together with the result from (Pacholski et al., 1997) leads to a 2-NExpTime algorithm. As for $C^2$, it is an open question whether this additional exponential blow-up is necessary. A positive answer would settle the same question for $C^2$ while a proof of the negative answer might give hints how the result for $C^2$ might be improved. For $\mathcal{ALCQ}$ with cardinality restrictions, we have partially solved this problem: with unary coding, the problem is ExpTime-complete whereas, for binary coding, it is NExpTime-hard.

## Acknowledgments

I would like to thank Franz Baader, Ulrike Sattler, and the anonymous referees for valuable comments and suggestions. This work was supported by the DFG, Project No. GR 1324/3–1.